\documentclass[10pt,twocolumn,letterpaper]{article}

\usepackage{subfigure}
\usepackage{siunitx}
\usepackage{cvpr}
\usepackage{times}
\usepackage{epsfig}
\usepackage{graphicx}
\usepackage{amsmath}
\usepackage{amssymb}

\usepackage{multirow}

\usepackage[breaklinks=true,bookmarks=false, colorlinks]{hyperref}

\cvprfinalcopy 

\def\cvprPaperID{1940} 

\ifcvprfinal\fi
\begin{document}

\title{POSEidon: Face-from-Depth for Driver Pose Estimation 
}

\author{Guido Borghi \quad Marco Venturelli \quad Roberto Vezzani \quad Rita Cucchiara\\
University of Modena and Reggio Emilia\\
{\tt\small \{name.surname\}@unimore.it}
 }


\maketitle
\thispagestyle{empty}

\begin{abstract}
Fast and accurate upper-body and head pose estimation is a key task for automatic monitoring of driver attention, a challenging context characterized by severe illumination changes, occlusions and extreme poses. 
In this work, we present a new deep learning framework for head localization and pose estimation on depth images.
The core of the proposal is a regressive neural network, called POSEidon, which is composed of three independent convolutional nets followed by a fusion layer, specially conceived for understanding the pose by depth. 
In addition, to recover the intrinsic value of face appearance for understanding head position and orientation, we propose a new Face-from-Depth model for learning image faces from depth. 
Results in face reconstruction are qualitatively impressive. 
We test the proposed framework on two public datasets, namely Biwi Kinect Head Pose and ICT-3DHP, and on Pandora, a new challenging dataset mainly inspired by the automotive setup.
Results show that our method overcomes all recent state-of-art works, running in real time at more than 30 frames per second.
\end{abstract}

\section{Introduction}
\label{sec:introduction}
\begin{figure}[t!]
    \centering
    \includegraphics[width=0.9\columnwidth]{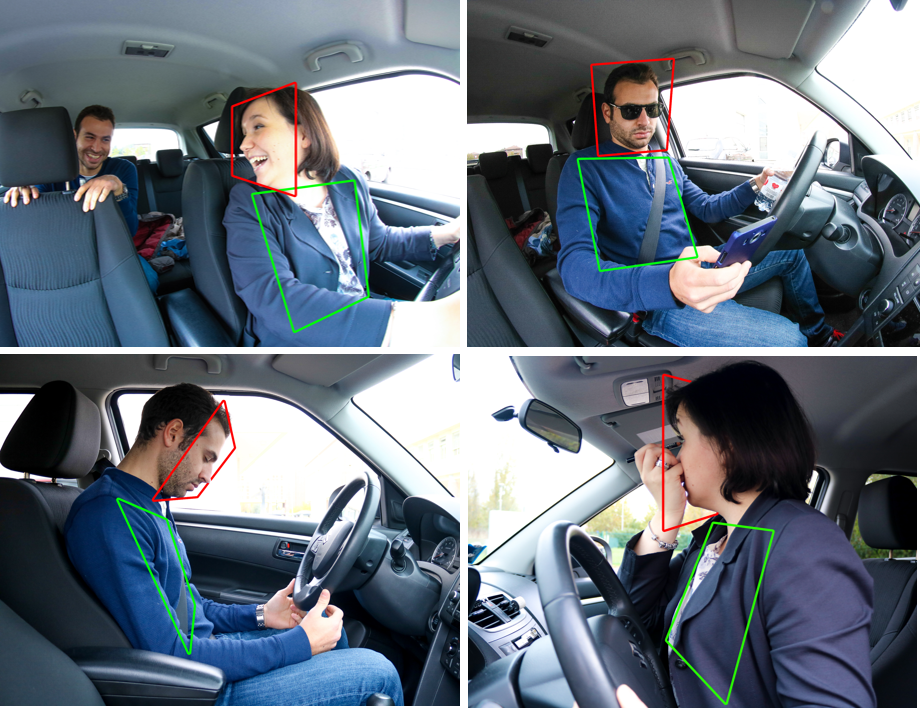}
    \caption{Some real situations in which head and upper-body pose estimation are useful to monitor driver's attention level: from the top-left, driver is talking with passengers, is playing with smartphone, is falling sleep and is looking at the rear-view mirror.
    }
    \label{fig:concept}
\end{figure}

\begin{figure}[t]
    \centering
    \includegraphics[width=0.99\columnwidth]{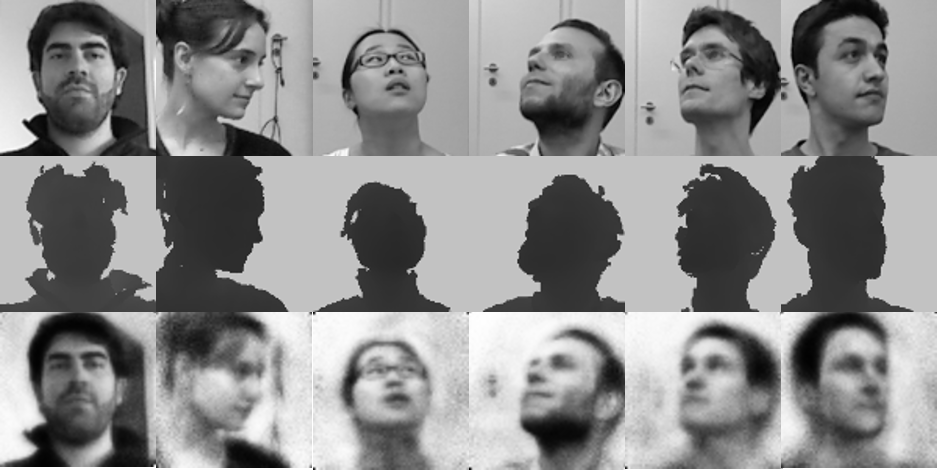}
    \caption{Examples of gray-level face images (bottom) reconstructed from the depth maps (middle). The corresponding ground truth is also shown (top). The first four subjects have been included in the training set, while the last two are completely new. 
    }
    \label{fig:facefromdepth}
\end{figure}

Nowadays, we are witnessing a revolution in the automotive field, where ICT technologies are becoming sometimes more important than the engine itself.\\ 
New solutions are required to solve many human-centered problems: semi-autonomous driving, driver behavior understanding, human-machine-interaction for entertainment, driver attention analysis for safe driving are just some examples. All of them, lay on the basic task of estimating driver pose, and in particular of the face and upper body parts, which are the mainly visible items of a driver.
Computer vision research \cite{Trivedi2009,tran2011,bergasa2006,doshi2012,czuprynski2014} achieved encouraging results, even if they are still not completely satisfactory due to some strong constraints of the context: reliability with strong pose changes, robustness to large occlusions (\textit{e.g.} glasses), in conjunction with non-intrusive capabilities, real time and low cost requirements (Fig. \ref{fig:concept}).
In addition, standard techniques based on intensity images are not always applicable, due to the poor illumination conditions during the night and the continuous illumination changes during the day. For this reasons, computer vision solutions based on illumination-insensitive data sources such as thermal \cite{trivedi2004occupant} or depth \cite{meyer2015} cameras are emerging.\\
Therefore, we propose a complete framework for driver monitoring based on depth images only, that can be easily acquired by commercial low-cost sensors placed inside the vehicles.
Starting from head localization, the ultimate goal of the framework is the estimation of the head and shoulder pose, measured as \textit{pitch}, \textit{roll} and \textit{yaw} rotation angles. 
To this aim, a new triple regressive Convolutional Neural Network architecture, called \textit{POSEidon}, is proposed, that combines depth, motion images and appearance. \\
One of the most innovative contribution is a \textit{Face-from-Depth} network, that is able to reconstruct gray-level faces directly from head depth images. This solution derives from the awareness that intensity face images are very useful to detect head pose \cite{ahn2014,drouard2015}: without having intensity data we would like to have similar benefits. Gray-level faces extracted by depth images have a qualitatively impressive similarity (Fig. \ref{fig:facefromdepth}). 
Summarizing, the novel contributions of the paper are the following: 
\begin{enumerate}
\item A complete and accurate framework, from head localization to head and shoulder pose estimation, based only on depth data, working in real time (30 fps);
\item A new \textit{Faces-from-Depth} architecture, to reconstruct gray-level face images directly from depth maps. To the best of our knowledge, this is the first proposal of this kind of approach;
\item A new dataset, called \textit{Pandora}, the first containing high resolution depth data with head and shoulder pose annotations.
\end{enumerate}

\section{Related Work}
\label{sec:related}
Head pose estimation approaches can rely on different input types: intensity images, depth maps, or both. 
In order to discuss related work, we adopt the classification proposed in \cite{meyer2015,fanelli2011}, updated and summarized in three main categories, namely  \textit{feature-based}, \textit{appearance-based} and \textit{3D model registration} approaches.

\textit{Feature-based} methods need facial (\textit{e.g.} nose, eyes) or pose-dependent features, that should be visible in all poses: consequently, these methods fail when features are not detected. In \cite{malassiotis2005} an accurate nose localization is used for head tracking and pose estimation on depth data.
Breitenstein \textit{et al.} \cite{breitenstein2008} used geometric features to identify nose candidates to produce the final pose estimation. HOG features \cite{dalal2005histograms} were extracted from RGB and depth images in \cite{yang2012,saeed2015}, then a Multi Layer Perceptron and a linear SVM were used for feature classification, respectively. Also \cite{vatahska2007,yang2002model,matsumoto2000} needed well visible facial features on RGB input images, and \cite{sun2008} on 3D data. \\
\indent \textit{Appearance-based} methods rely on one or more classifiers that use raw input images, trained to perform head pose estimation. In \cite{Seemann2004} RGB and depth data were combined, exploiting a neural network to perform head pose prediction. Fanelli \textit{et al.} \cite{fanelli2011,fanelli2011dagm,fanelli2013} trained Random Regression Forest for both head detection and pose estimation on depth images. Tulyakov \textit{et al.} \cite{tulyakov2014} used a cascade of tree classifiers to tackle extreme head pose estimation task. A Convolutional Neural Network (CNN) based on RGB input images is exploited in \cite{ahn2014}.
Recently, in \cite{mukherjee2015} a multimodal CNN was proposed to estimate gaze direction: a regression approach was only approximated through a 360-classes classifier. 
Synthetic datasets were used to train CNNs, that generally require a huge amount of data, \textit{e.g.} \cite{liu3d2016}.\\
\indent \textit{3D model registration} approaches create a head model from the acquired data; frequently, a manual initialization is required. In \cite{papazov2015} facial point clouds were matched with pose candidates, through a triangular surface patch descriptor. In \cite{baltruvsaitis2012} intensity and depth data were used to build a 3D constrained local method for robust facial feature tracking. In \cite{ghiass2015} a 3D morphable model is fitted, using both RGB and depth data to predict head pose. Also \cite{baltruvsaitis2012,cai2010,blanz1999,cao2013,bleiweiss2010,rekik2013} built 3D facial model for head tracking, animation and pose estimation.\\
\indent Remaining methods regard head pose estimation task as an \textit{optimization} problem: \cite{padeleris2012} used the Particle Swarm Optimization (PSO) \cite{kennedy2011}; \cite{bar2012} exploited the Iterative Closest Point algorithm (ICP) \cite{low2004}; \cite{meyer2015} combined PSO and ICP techniques. \cite{kondori2011} used a least-square technique to minimize the difference between the input depth change rate and the prediction rate. 
Besides, other works use linear or nonlinear regression with extremely low resolution images \cite{chen2016}. HOG features and a Gaussian locally-linear mapping model were used in \cite{drouard2015}. Finally, recent works produce head pose estimations performing face alignment task \cite{zhu2015}.\\
\begin{figure*}[th!]
    \centering
    \includegraphics[width=0.9\linewidth]{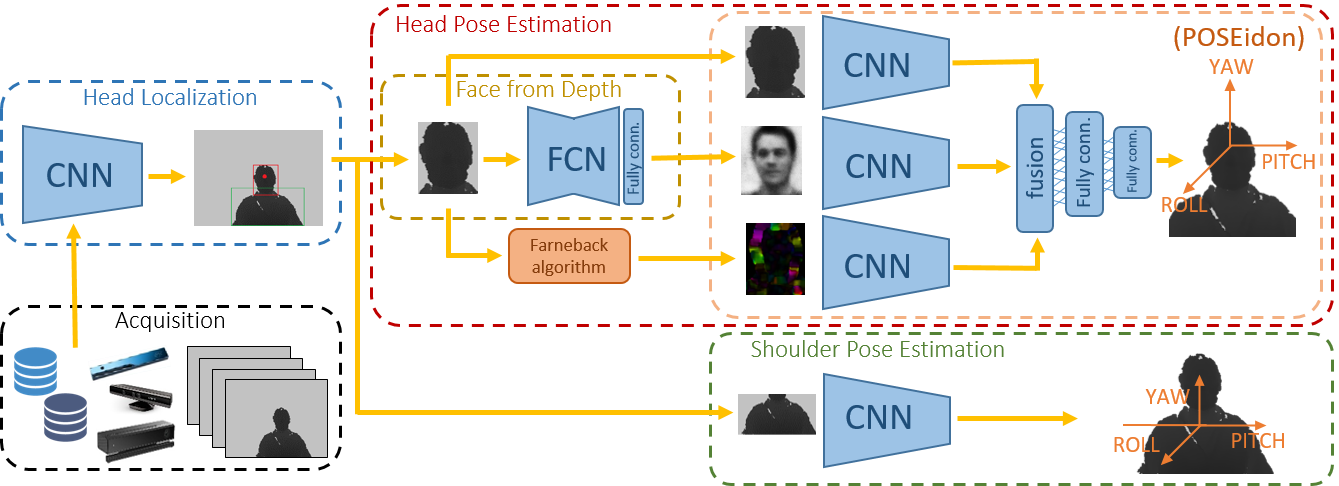}
    \caption{Overview of the whole \textit{POSEidon} framework. Depth input images are acquired by low cost sensors (black) and provided to a head localization CNN (blue) to suitably crop the images around the upper-body or head regions. The first is exploited by the shoulder pose estimation task (green), while the second is selected for the head pose estimation (red) obtained through the \textit{POSEidon} network (orange). In the center, the \textit{Face-from-Depth} net (yellow) which produces gray-level images of the face from the depth map. [best in color]}
    \label{fig:general}
\end{figure*}
\indent Several works based on head pose estimation do not take in consideration head localization task.
To propose a complete head pose estimation framework, it is necessary to perform a head detection, finding the complete head or a particular point, for example the head center. With RGB images Viola and Jones \cite{viola2004} face detector is often exploited, \textit{e.g.} in \cite{ghiass2015,cai2010,rekik2013,baltruvsaitis2012,Seemann2004}. A different approach demands the head location to a classifier, \textit{e.g.} \cite{fanelli2011,tulyakov2014}. As reported in \cite{meyer2015}, these approaches suffer due to the lack of generalization capabilities with different acquisition devices. \\
\indent Only few works in literature tackle the problem of driver body pose estimation focusing only on upper-body part or in automotive context. Ito \textit{et al.} \cite{ito2008} adopting an intrusive approach, placed six marker points on driver body to predict some typical driving operations. A 2D driver body tracking system was proposed in \cite{datta2008}, but a manual initialization of the tracking model is strictly required. In \cite{trivedi2004occupant} a thermal long-wavelength infrared video camera was used to analyze occupant position and posture. In \cite{tran2009introducing} an approach for upper body tracking system using 3D head and hands movements was developed.

\section{The POSEidon framework} 
\label{pipeline}
An overview of the \textit{POSEidon} framework is depicted in Figure \ref{fig:general}. The final goal is the estimation of the pose of the driver's head and shoulders, defined as the mass center position and the corresponding orientation relatively to the reference frame of the acquisition device \cite{Trivedi2009}. The orientation is represented using three \textit{pitch}, \textit{roll} and \textit{yaw} rotation angles. \textit{POSEidon} directly processes the stream of depth frames captured in real time by a commercial sensor (\textit{e.g.}, \textit{Microsoft Kinect}). 
Position and size of the head in the foreground are estimated by a head localization module based on a regressive CNN (Sect. \ref{sec:headlocalization}). The output provided is used to crop the input frames around the head or the shoulder bounding boxes, depending on the further pipeline type. Frames cropped around the head are fed to the head pose estimation block, while the others are exploited to estimate the shoulders pose.
The core components of the system are the \textit{Face-from-Depth} network (Sect.~\ref{sec:depth-to-gray}), and \textit{POSEidon} (Sect.~\ref{sec:poseidon}), the network which gives the name to the whole framework. Its trident shape is due to the three included CNNs, each working on likewise sources: depth, \textit{Face-from-Depth} and motion images data. The first one -- \textit{i.e.}, the CNN directly working on depth data -- plays the main role on the pose estimation, while the other two cooperate to reduce the estimation error. \\

\section{Face-from-Depth network} 
\label{sec:depth-to-gray}

\begin{figure}[t!]
    \centering
    \includegraphics[width=\columnwidth]{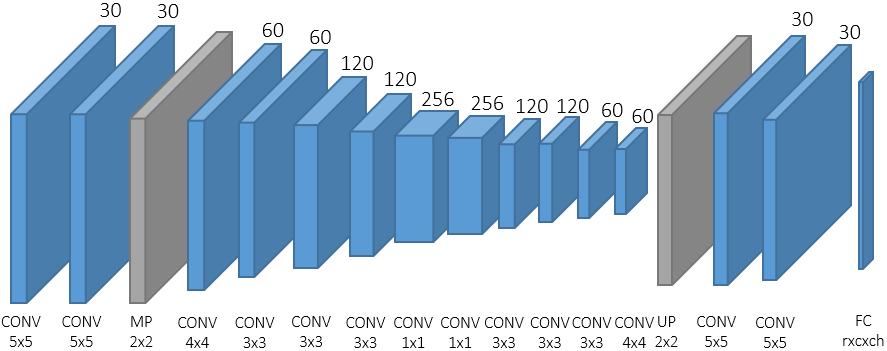}
    \caption{Architecture of the \textit{Face-from-Depth} network.}
    \label{fig:ae_network}
\end{figure}

\textit{Face-from-Depth} (FfD) is one of the most innovative elements of the framework. Due to illumination issues, the appearance of the face is not always available in many scenarios, \textit{e.g.} inside a vehicle. On the contrary, depth maps are invariant to illumination conditions but lacks of texture details. We aim to investigating if it is possible to imagine the appearance of a face given the corresponding depth data. 
The \textit{Face-from-Depth} network has been created to this goal, even if the output is not always realistic and visually pleasant: however, the promising results confirm their positive contribution in the estimation of the head pose.\\
The proposed architecture fuses the key aspects of autoencoders  \cite{masci2011stacked} and fully convolutional \cite{long2015fully} neural networks: it is composed by 14 convolutional layers, plus a fully connected layer at the end (Fig. \ref{fig:ae_network}). A single $2 \times 2$ max-pooling layer has been inserted after the second layer, and a corresponding up-sampling layer after the thirteenth. Besides, two zero-padding layers are added after the first and the second convolutional layers, respectively. 
We train the network in a single stage, with input head images resized to $64 \times 64$ pixels. The hyperbolic tangent activation function is used and best training performances are reached through the self adaptive \textit{Adadelta} optimizer \cite{zeiler2012adadelta}. A specific loss function is exploited to highlight the central area of the image, where the face is supposed to be after the cropping step, and takes in account the distance between the reconstructed image and the corresponding gray-level ground truth:
\begin{equation}
L = \frac{1}{R\cdot C} \sum_i^R \sum_j^C \left( || y_{ij} - \bar{y}_{ij} || ^ 2 _2 \cdot w_{ij}^{\mathcal{N}} \right)
\end{equation}
\noindent where $R,C$ are the number of rows and columns of the input images, respectively. $y_{ij}, \bar{y}_{ij} \in \mathcal{R}^{ch}$ are the intensity values from ground truth ($ch=1$) and predicted appearance images. Finally, the term $w_{ij}^{\mathcal{N}}$ introduces a bivariate Gaussian prior mask. Best results have been obtained using $\mu=[\frac{R}{2},\frac{C}{2}]^T$ and $\Sigma = \mathbb{I} \cdot [ \left(R / \alpha \right)^2, \left( C / \beta \right)^2 ]^T$ with $\alpha$ and $\beta$ empirically set to $3.5, 2.5$ for squared images of $R=C=64$.
Some visual examples of input, output and ground-truth images are reported in Figure \ref{fig:facefromdepth}.

\section{Pose Estimation from depth}
\label{sec:poseEstimation}
\subsection{Head Localization Network} 
\label{sec:headlocalization}
\begin{figure}[b!]
    \centering
    \includegraphics[width=\columnwidth]{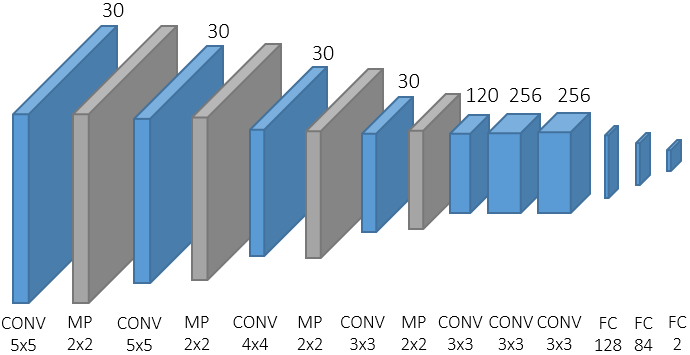}
    \caption{Architecture of the head localization network.}
    \label{fig:networklocalization}
\end{figure}

In this step we design a network to perform head localization, relying on the main assumption that a single person is in the foreground. The desired network outputs are the image coordinates $(x_H,y_H)$ of the head center, or rather, the average position of all head points in the frame \cite{shotton2013}.
Details on the deep architecture adopted are reported in Figure \ref{fig:networklocalization}. A limited depth and small sized filters have been chosen to meet real time constraint while keeping satisfactory performance. For the same reason, input images are firstly resized to $160 \times 132$ pixels. A max-pooling layer is run after each of the first four convolutional layers, while a dropout regularization ($\sigma=0.5$) is exploited in fully connected layers. The hyperbolic tangent activation (\textit{tanh}) function is adopted, in order to map continuous output values to a predefined range $[-\infty, +\infty] \rightarrow [-1, +1]$. The network has been trained by \textit{Stochastic Gradient Descent} (SGD) \cite{krizhevsky2012} and the $L_2$ loss function.\\ 
Given the head position $(x_H,y_H)$ in the frame, a dynamic size algorithm provides the head bounding box with barycenter $(x_H, y_H)$ and width $w_H$ and height $h_H$, around which the frames are cropped:
\begin{equation}
w_H=\frac{f_x \cdot R_x}{D}, \quad h_H=\frac{f_y \cdot R_y}{D}
\label{eq:headBB}
\end{equation}
\noindent where $f_x, f_y$ are the horizontal and the vertical focal lengths in pixels of the acquisition device, respectively. $R_x, R_y$ are the average width and height of a face (for head pose task $R_x=R_y=320$) and $D$ is the distance between the head center and the acquisition device, computed averaging the depth values around the head center. \\
Some examples of the bounding boxes estimated by the network are superimposed in Figure \ref{fig:final demo}.

\subsection{POSEidon} \label{sec:poseidon}
The \textit{POSEidon} network is mainly obtained as a fusion of three CNNs and has been developed to perform a regression on the 3D pose angles. As a result, continuous Euler values -- corresponding to the \textit{yaw}, \textit{pitch} and \textit{roll} angles -- are estimated (right part of Fig. \ref{fig:general}). The three \textit{POSEidon} components have the same shallow architecture based on 5 convolutional layers with kernel size of $5 \times 5$, $4 \times 4$ and $3 \times 3$, max-pooling is conducted only on the first three layers. The first four convolutional layers have 32 filters each, the last one has 128 filters. At the end of the network, there are 3 fully connected layers, with 128, 84 and 3 neurons, respectively. Also in this case \textit{tanh} function is exploited: we are aware that \textit{ReLU} \cite{nair2010rectified} converges faster, but we obtain better performance in term of accuracy prediction.  The three networks are fed with different input data types: the first one, directly takes as input the head-cropped depth images; the second one is connected to the \textit{Face-from-Depth} output and the last one operates on motion images, obtained applying the standard \textit{Farneback} algorithm \cite{farneback2001} on pairs of consecutive depth frames. 
A fusion step combines the contributions of the three above described networks: in this case, the last fully connected layer of each component is removed. 
Different fusion approaches that have been proposed \cite{park2016} are investigated. Given two feature maps $x^a, x^b$ with a certain width $w$ and height $h$, for every feature channel $d^x_a, d^x_b$ and $y \in R^{w \times h \times d}$:
\begin{itemize}
\item \textbf{Multiplication}: computes the element-wise product of two feature maps, as
$ y^{mul}=x^a \circ x^b, d^y = d^x_a = d^x_b $
\item \textbf{Concatenation}: stacks two features maps, without any blend
$y^{cat} = [x^a | x^b], d^y = d^x_a + d^x_b$
\item \textbf{Convolution}: stacks and convolves feature maps with a filter $k$ of size $1 \times 1 \times (d^x_a + d^x_b)/2$ and $\beta$ as bias term, 
$ y^{conv} = y^{cat} \ast k + \beta, \quad d^y = (d^x_a+d^x_b) / 2 $
\end{itemize}
%
Final \textit{POSEidon} framework exploits a combination of two fusing methods, in particular a convolution followed by a concatenation. After the fusion step, three fully connected layers composed of 128, 84 and 3 activations respectively and two dropout regularization ($\sigma=0.5$) complete the architecture.
\textit{POSEidon} is trained with a double-step procedure. First, each individual network is trained with the following $L_2^w$ weighted loss:
\begin{equation}
L_2^w = \sum_{i=1}^3 \big \lVert w_i \cdot \left(y_i - f(x_i)\right) \big \rVert _2
\label{eq:l2_pesata}
\end{equation}
where $ w_i \in [0.2, 0.35, 0.45]$: this weight distribution gives more importance to the yaw angle, which is preponderant in the selected automotive context. During the individual training step, the last fully connected layer of each network is preserved, then is removed to perform the second training phase: holding the weights learned for the trident components, the new training phase is carried out on the last three fully connected layers of \textit{POSEidon}, with the loss function $L^w_2$ reported in Equation \ref{eq:l2_pesata}. 
In all training steps, the SGD optimizer \cite{krizhevsky2012} is exploited, the learning rate is set initially to $10^{-1}$ and then is reduced by a factor 2 every 15 epochs.


\begin{table*}[h]
\centering
\small
\begin{tabular}{ccccccc}
\multicolumn{7}{c}{\textsc{Head Pose estimation error [euler angles]}}\\
\hline
\textbf{Method} &\textbf{Year} &\textbf{Data}  &\textbf{Pitch} &\textbf{Roll}  &\textbf{Yaw}    &\textbf{Avg}     \\ 
\hline \hline
Fanelli \cite{fanelli2011}  &2011          & Depth 				& 8.5 $\pm$ 9.9			   & 7.9 $\pm$ 8.3			& 8.9 $\pm$ 13.0	&8.43 $\pm$ 10.4   \\ \hline
Yang \cite{yang2012}			&2012	  & RGB + Depth			& 9.1 $\pm$ 7.4			   & 7.4 $\pm$ 4.9			& 8.9 $\pm$ 8.3	&8.5 $\pm$ 6.9 	 \\ \hline
Padeleris \cite{padeleris2012}	&2012	  & Depth				&6.6   & 6.7	& 11.1	&8.1 	 \\ \hline
Rekik \cite{rekik2013}		&2013	  &RGB + Depth			&4.3   & 5.2	& 5.1   &4.9  \\ \hline
Baltrusaitis \cite{baltruvsaitis2012}  &2012    & RGB + Depth	& 5.1 					   & 11.3					& 6.3	&7.6 \\ \hline
Ahn \cite{ahn2014}*      &2014        	 & RGB 					& 3.4 $\pm$ 2.9			   & 2.6 $\pm$ 2.5			& 2.8 $\pm$ 2.4		&2.9 $\pm$ 2.6  \\ \hline
Martin \cite{Martin2014}*      &2014        	 & Depth		& 2.5 		  			   & 2.6 	  				& 3.6		 &2.9 \\ \hline
Saeed \cite{saeed2015}      &2015        & RGB + Depth 			& 5.0 $\pm$ 5.8			   & 4.3 $\pm$ 4.6			& 3.9 $\pm$ 4.2		&4.4 $\pm$ 4.9  \\ \hline
Papazov \cite{papazov2015}  &2015  	  	 & Depth				& 2.5 $\pm$ 7.4 		   & 3.8 $\pm$ 16.0			& 3.0 $\pm$ 9.6 	&4.0 $\pm$ 11.0  \\ \hline
Drouard \cite{drouard2015}  &2015  	  	 & RGB					& 5.9 $\pm$ 4.8 		   & 4.7 $\pm$ 4.6			& 4.9 $\pm$ 4.1 	 &5.2 $\pm$ 4.5 \\ \hline
Meyer \cite{meyer2015}  &2015  	  	 	 & Depth				& 2.4 		   			   & 2.1			& 2.1 	 &2.2 \\ \hline
Liu \cite{liu3d2016}  &2016  	  	 	 & RGB					& 6.0 $\pm$ 5.8 		   & 5.7 $\pm$ 7.3			& 6.1 $\pm$ 5.2 	 &5.9 $\pm$ 6.1 \\ \hline
\textbf{POSEidon}   &2016  	  	 	 & Depth				& \textbf{1.6} $\pm$ \textbf{1.7}  & \textbf{1.8} $\pm$ \textbf{1.8}  	   			& \textbf{1.7} $\pm$ \textbf{1.5}  	&\textbf{1.7} $\pm$ \textbf{1.7}  \\ \hline
\end{tabular}
\caption{Results on \textit{Biwi} dataset. Input cropping is done using the ground truth head position.}
\label{tab:resBiwiPose}
\end{table*}

\subsection{Shoulder pose estimation network} \label{sec:shoulders}
The framework is completed with an additional network for the estimation of the shoulder pose. We employ the same architecture adopted for the head (Section \ref{sec:poseidon}), performing regression on the same three pose angles. Starting from the head center position (Section \ref{sec:headlocalization}), the depth input images are crop around the driver neck, using a bounding box $\{x_S, y_S, w_S, h_S\}$ with barycenter $(x_S = x_H, y_S = y_H - (h_H/4))$, and width and height obtained as described in Equation \ref{eq:headBB}, but with different values of $R_x, R_y$ to produce a rectangular crop: these values are tested and discussed in Section \ref{sec:results}.
The network is trained with SGD optimizer \cite{krizhevsky2012}, using the weighted $L_2^w$ loss function described above (see Eq. \ref{eq:l2_pesata}). Hyperbolic tangent is exploited as activation functions as usual.

\section{Datasets}
\label{sec:dataset}
Network training and testing phases have been done exploiting two publicly available datasets, namely \textit{Biwi Kinect Head Pose} and \textit{ICT-3DHP}. In addition, we collect a new dataset, called \textit{Pandora}, which also contains shoulder pose annotations. Data augmentation techniques are employed to enlarge the training set, in order to achieve space invariance and avoid over fitting \cite{krizhevsky2012}. Random translations on vertical, horizontal and diagonal directions, jittering, zoom-in and zoom-out transformation of the original images have been exploited. Percentile-based contrast stretching, normalization and scaling of the input images are also applied to produce zero mean and unit variance data. \\
Follows a detailed description of the three adopted datasets.

\subsection{Biwi Kinect Head Pose dataset}
Fanelli \textit{et al.} \cite{fanelli2011} introduced this dataset in 2013. It is acquired with the \textit{Microsoft Kinect} sensor, a structured IR light device. It contains about 15k frame, with RGB ($640 \times 480$) and depth maps ($640 \times 480$). Twenty subjects have been involved in the recordings: four of them  were recorded twice, for a total of 24 sequences. The ground truth of yaw, pitch and roll angles is reported together with the head center and the calibration matrix. The original paper does not report the adopted split between training and testing sets; fair comparisons are thus not guarantee. To avoid this, we clearly report the adopted split in the following.

\subsection{ICT-3DHP dataset}
\textit{ICT-3DHP} dataset has been introduced by Baltrusaitis \textit{et al.} in 2012 \cite{baltruvsaitis2012}. It is collected using the \textit{Microsoft Kinect} sensor and contains RGB images and depth maps of about 14k frames, divided in 10 sequences. The image resolution is $640 \times 480$ pixels. An hardware sensor (\textit{Polhemus Fastrack}) is exploited to generate the ground truth annotation. The device is placed on a white cap worn by each subject, visible in both RGB and depth frames. Moreover, the presence of few subjects and the limited number of frames make this dataset unsuitable for deep learning approaches.

\begin{figure}[t!]
    \centering
    \includegraphics[width=\columnwidth]{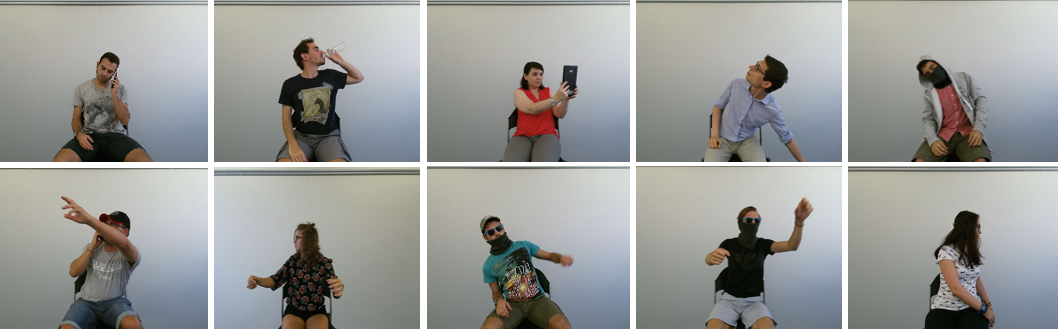}
    \caption{Sample frames from the \textit{Pandora} dataset.}
    \label{fig:pandora}
\end{figure}

\subsection{Pandora dataset}
We collect a new challenging dataset, called \textit{Pandora}. 
The dataset has been specifically created for the tasks described in the paper (\textit{i.e.}, head center localization, head pose and shoulder pose estimation) and is inspired by the automotive context. A frontal fixed device acquires the upper body part of the subjects, simulating the point of view of camera placed inside the dashboard. Among the others, the subjects also perform driving-like actions, such as grasping the steering wheel, looking to the rear-view or lateral mirrors, shifting gears and so on. \textit{Pandora} contains 110 annotated sequences using 10 male and 12 female actors. Each subject has been recorded five times.  

\textit{Pandora} is the first publicly available dataset which combines the following features:
\begin{itemize}
\item \textbf{Shoulder angles}: in addition to the head pose annotation, \textit{Pandora} contains the ground truth data of the shoulder pose expressed as yaw, pitch and roll. 
\item \textbf{Wide angle ranges}: subjects perform wide head ($\pm \ang{70}$ roll, $\pm \ang{100}$ pitch and $\pm \ang{125}$ yaw) and shoulder ($\pm \ang{70}$ roll, $\pm \ang{60}$ pitch and $\pm \ang{60}$ yaw) movements. For each subject, two sequences are performed with constrained movements, changing the yaw, pitch and roll angles separately, while three additional sequences are completely unconstrained.
\item \textbf{Challenging camouflage}: garments as well as various objects are worn or used by the subjects to create head and/or shoulder occlusions. For example, people wear prescription glasses, sun glasses, scarves, caps, and  manipulate smartphones, tablets or plastic bottles. 
\item \textbf{Deep-learning oriented}: the dataset contains more than 250k full resolution RGB ($1920 \times 1080$) and depth images ($512 \times 424$) with the corresponding annotation.
\item \textbf{Time-of-Flight (ToF) data}: a \textit{Microsoft Kinect One} device is used to acquire depth data, with a better quality than other datasets created with the first \textit{Kinect} version \cite{sarbolandi2015}. 
\end{itemize}

\noindent Each frame of the dataset is composed of the RGB appearance image, the corresponding depth map, the 3D coordinates of the skeleton joints corresponding to the upper body part, including the head center and the shoulder positions. For convenience's sake, the 2D coordinates of the joints on both color and depth frames are provided as well as the head and shoulder pose angles with respect to the camera reference frame. Shoulder angles are obtained through the conversion to Euler angles of a corresponding rotation matrix, obtained from a user-centered system \cite{papadopoulos2014} and defined by the following unit vectors $(N_1,N_2,N_3)$:
\begin{equation}
\begin{array}{cc}
N_1=\frac{p_{RS}-p_{LS}}{\| p_{RS}-p_{LS} \|} & U=\frac{p_{RS}-p_{SB}}{\|p_{RS}-p_{SB}\|}\\ \\
N_3=\frac{N_1 \times U}{\| N_1 \times U \|} & N_2=N_{1} \times N_{3}
\end{array}
\label{eq:refframe}
\end{equation}
where $p_{LS}$, $p_{RS}$ and $p_{SB}$ are the 3D coordinates of the left shoulder, right shoulder and spine base joints.
The annotation of the head pose angles has been collected using a wearable \textit{Inertial Measurement Unit} (IMU) sensor. 
The sensor has been worn by the subjects in a non visible position (\textit{i.e.}, on the rear of the head) to avoid  distracting artifacts on both color and depth images. IMU sensor has been calibrated and aligned at the beginning of each sequence, assuring the reliability of the provided angles. The dataset is publicly available (\url{http://imagelab.ing.unimore.it/pandora/}).

\begin{table*}[th!]
\centering
\small
\begin{tabular}{c c c ccccc}
\multicolumn{8}{c}{\textsc{Head Pose estimation error [euler angles]}}\\
\hline
\textbf{Architecture} &
\textbf{Input} &\textbf{Cropping}&\textbf{Fusion} &\multicolumn{3}{c}{\textbf{Head}}   &\textbf{Accuracy} \\
 	&	& &			 &Pitch					&Roll			  &Yaw		   	  	 &	\\ \hline \hline
\multirow{5}{*}{Single CNN} & depth & 	&-			 &8.1 $\pm$ 7.1 		&6.2 $\pm$ 6.3		&11.7 $\pm$ 12.2 	 &0.553	\\ \cline{2-8}
&depth	& $\surd$ 			&-			 &6.5 $\pm$ 6.6 		&5.4 $\pm$ 5.1		&10.4 $\pm$ 11.8 	 &0.646	\\ \cline{2-8}
&FfD   & $\surd$   		&-			 &6.8 $\pm$ 7.0 		&5.7 $\pm$ 5.7		&10.5 $\pm$ 14.6 	 &0.647	\\ \cline{2-8}
&gray-level 	& $\surd$			&-			 &7.1 $\pm$ 6.6 		&5.6 $\pm$ 5.8		&9.0 $\pm$ 10.9 	 &0.639	\\ \cline{2-8}
&MI& $\surd$		&-			 &7.7 $\pm$ 7.5 		&5.3 $\pm$ 5.7		&10.0 $\pm$ 12.5 	 &0.609	\\ \hline
\multirow{2}{*}{Double CNN} & depth + FfD	& $\surd$	&concat		 &5.6 $\pm$ 5.0 		&4.9 $\pm$ 5.0		&9.8  $\pm$ 13.4	 &0.698	\\ \cline{2-8}
& depth + MI	& $\surd$		&concat		 &6.0 $\pm$ 6.1 		&4.5 $\pm$ 4.8		&9.2 $\pm$11.5       &0.690	\\ \hline
\multirow{3}{*}{POSEidon} & depth + FfD + MI 	& $\surd$		&concat 	 &6.3 $\pm$ 6.1			&5.0 $\pm$ 5.0      &10.6 $\pm$14.2      &0.657	\\ \cline{2-8}
& depth + FfD + MI 	& $\surd$		&mul+concat  &5.6 $\pm$ 5.6   		&4.9 $\pm$ 5.2 	    &9.1$\pm$ 11.9    	 &0.712	\\ \cline{2-8}
& depth + FfD + MI 	& $\surd$	&conv+concat &5.7 $\pm$ 5.6 		&4.9 $\pm$ 5.1		&9.0 $\pm$ 11.9       &0.715	\\ \hline
\end{tabular}

\caption{Results of the head pose estimation on \textit{Pandora} comparing different system architectures. The baseline is a single CNN working on the source depth map. The accuracy is the percentage of correct estimations ($err<$\ang{15}). FfD: Face-from-Depth, MI: Motion Images.}
\label{tab:resOurDataset}
\end{table*}

\begin{figure}[h!]
\centering
\subfigure[left]{\includegraphics[width=0.16\columnwidth]{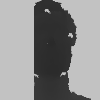}\label{latleft}} 
\subfigure[top]{\includegraphics[width=0.16\columnwidth]{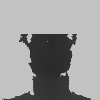}\label{top}} 
\subfigure[right]{\includegraphics[width=0.16\columnwidth]{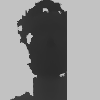}\label{latright}} 
\subfigure[bottom]{\includegraphics[width=0.16\columnwidth]{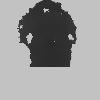}\label{down}} 
\subfigure[middle]{\includegraphics[width=0.16\columnwidth]{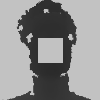}\label{center}} 
\caption{Visual examples of the simulated occlusion types applied on a \textit{Biwi} frame.} 
\label{fig:occlusion} 
\end{figure}
 
\section{Experimental results} \label{sec:results}

\begin{figure}[t!]
    \centering
    \includegraphics[width=1\columnwidth]{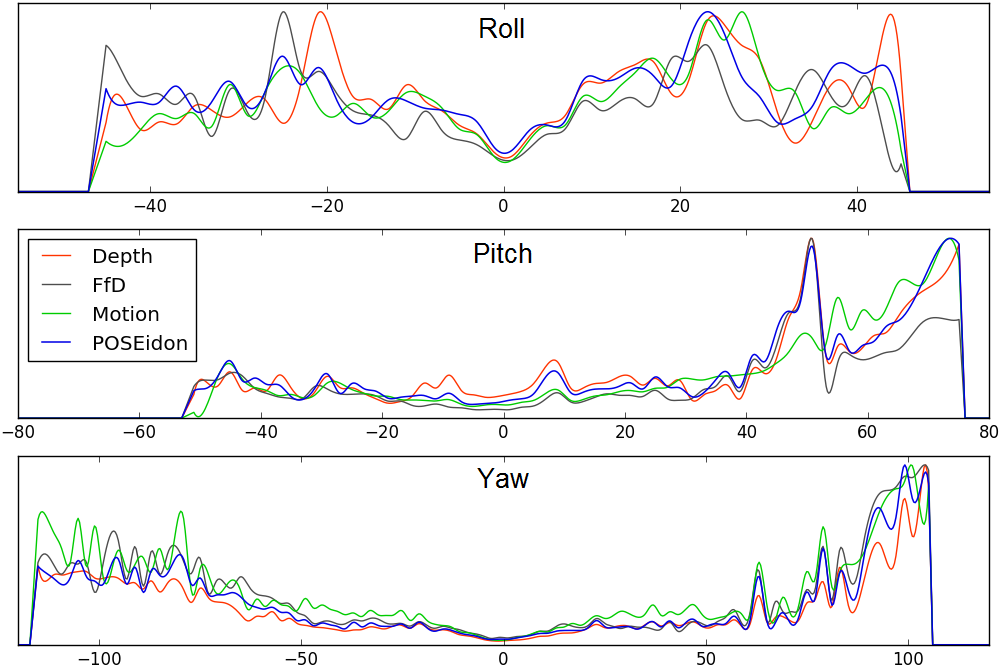}
    \caption{Error distribution of each \textit{POSEidon} components on \textit{Pandora} dataset. On x-axis are reported the ground truth angles, on y-axis the distribution of error for each input type.
}
    \label{fig:erroriperangolo}
\end{figure}

\begin{figure}[b!]
    \centering
    \includegraphics[width=0.87\columnwidth]{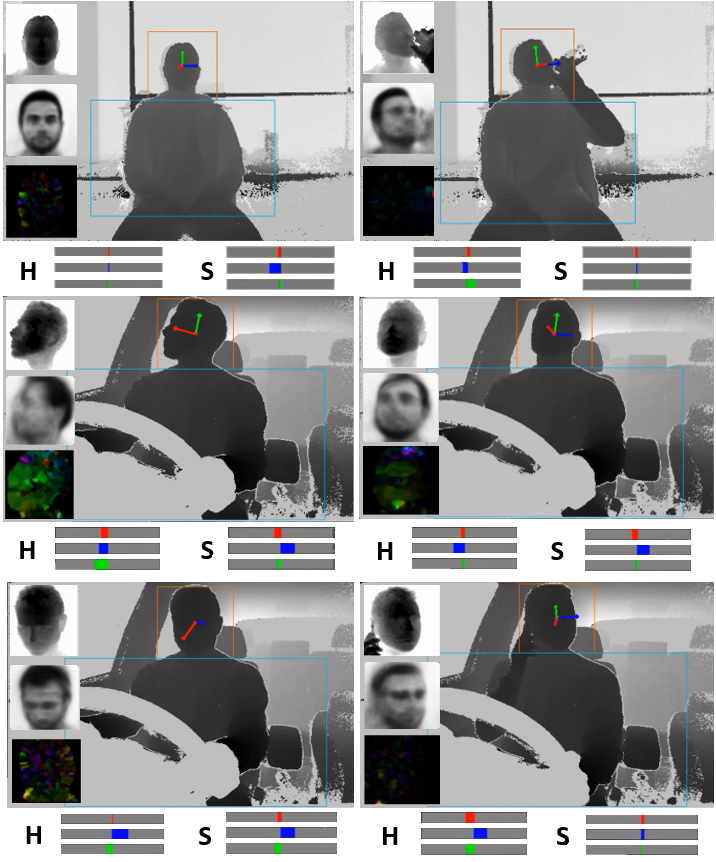}
    \caption{Visual examples of the proposed framework output. Head (H) and shoulder (S) pose angles are reported as bars centered at \ang{0}. Depth maps, \textit{Face-from-Depth} and motion image inputs are depicted on the left of each frame. [best in colors]}
    \label{fig:final demo}
\end{figure}

The proposed framework has been deeply tested using dataset described in Section \ref{sec:dataset}. 
\begin{table}[b!]
\centering
\small
\begin{tabular}{cccc}
\hline
\textbf{Occluded part}  &\multicolumn{3}{c}{\textbf{Head}}   \\
 						&Pitch		&Roll			&Yaw    			\\ \hline \hline
(a) left	&2.6$ \pm$ 3.0		&4.0 $\pm$ 2.9    &7.8 $\pm$ 8.1  		\\ \hline
(b)	top 	&42.5 $\pm$ 21.2 	&12.3 $\pm$ 9.3   &10.2 $\pm$ 7.6 		\\ \hline
(c) right	&2.1 $\pm$ 1.8		&2.8 $\pm$ 2.6    &8.4 $\pm$ 8.5 		\\ \hline
(d) bottom		&4.2 $\pm$ 3.3		&4.3 $\pm$ 3.5    &4.0 $\pm$ 3.0 		\\ \hline
(e) middle	&11.0 $\pm$ 5.3		&3.0 $\pm$ 2.8    &6.1 $\pm$ 4.9  		\\ \hline
random		 	&12.5 $\pm$ 18.3	&5.3 $\pm$ 6.1    &7.4 $\pm$ 7.1  		\\ \hline
\end{tabular}
\caption{Estimation errors of \textit{POSEidon} in presence of simulated occlusions. The system is fed with images from the \textit{Biwi} dataset occluded using the masks illustrated in figure \ref{fig:occlusion}. Results of the last line are obtained by applying a random mask to each frame.}
\label{tab:occlusions}
\end{table}
\noindent Besides, an ablation study has been evaluated on \textit{Pandora}.\\ 
Sequences of subjects 10, 14, 16 and 20 have been used for testing, the remaining for training. Table \ref{tab:resOurDataset} reports an internal evaluation, providing mean and standard deviation of the estimation errors obtained on each angle and for each system configuration. Similar to Fanelli \textit{et al.} \cite{fanelli2011}, we also report the mean accuracy as percentage of good estimations (\textit{i.e.}, angle error below \ang{15}).
The first line of Table \ref{tab:resOurDataset} shows the performance of a baseline system, obtained using the pose estimation network only and input depth frames are directly fed to the network without processing and crop. 
The crop step is included instead in the configurations of the other rows, using the ground truth head position as center. Results obtained using single networks, couples of them and the complete \textit{POSEidon} architecture are shown. The last row highlights the best performance reached using \textit{conv} fusion of couples of input types, followed by the \textit{concat} step. Even if the choice of the fusion method has a limited effect (as deeply investigated in \cite{park2016,feichtenhofer2016}), the most significant improvement of the system is reached exploiting the three input types together. \\
Figure \ref{fig:erroriperangolo} shows a comparison of the estimation errors made by each trident component: each graph plots the error distribution of a specific network configuration with respect to the ground truth value. Depth data allows to reach the lowest error rates for frontal heads, while the other input data types are better in presence of rotated poses. The graphs highlight the averaging capabilities of \textit{POSEidon} too. 
\\
Table \ref{tab:resOurDataset} includes an indirect evaluation of the reconstruction capabilities of the \textit{Face-from-Depth} network. The results reported on the third and fourth rows are obtained using the network described in Section \ref{sec:poseidon} with the reconstructed appearance image and the original gray-level images as input, respectively. The similar results confirm that the obtained image reconstruction is sufficiently accurate, at least for the pose estimation task. 
We compared the results of \textit{POSEidon} with state-of-art, using the \textit{Biwi} dataset. According with Fanelli \textit{et al.} \cite{fanelli2011}, 18 subjects are used to train the system while two for the test. More specifically, we exploited the sequences 11 and 12 for testing and the remaining for training. Table \ref{tab:resBiwiPose} reports the corresponding results as indicated in the cited papers.
\textit{POSEidon} achieves impressive results on \textit{Biwi} dataset: the mean error is under \ang{2} for all of the three angles, with a small standard deviation. The system overcomes all the reported methods, included the recent proposal by Meyer \textit{et al.} \cite{meyer2015}. The performance are better than other approaches based on deep learning, 3D data and regression \cite{ahn2014,mukherjee2015}. Moreover, \textit{POSEidon} also overcomes the approaches working on appearance data.
The proposals marked with a star (*) do not follow the same split or apply a different testing procedure: 
thus, the comparison with them may not be fair. Results of \cite{padeleris2012} reported in table have been taken from \cite{meyer2015} for the sake of comparability.\\
\begin{table}[b!]
\centering
\small
\begin{tabular}{cc|cccc}
\hline
\multicolumn{2}{c}{\textbf{Parameters}}  &\multicolumn{3}{c}{\textbf{Shoulders}}   & \textbf{Accuracy}\\
$R_x$  &$R_y$ 		&Pitch		&Roll			&Yaw    &		\\ \hline \hline
\multicolumn{2}{c|}{No crop} &2.5 $\pm$ 2.3			&3.0 $\pm$2.6    &3.7 $\pm$ 3.4 &0.877				\\ \hline
700	 	&250		&2.9 $\pm$ 2.6			&2.6 $\pm$2.5    &4.0 $\pm$ 4.0 	 	&0.845 	\\ \hline
850	 	&250		&2.4 $\pm$ 2.2			&2.5 $\pm$2.2    &3.1 $\pm$ 3.1 	 	&0.911 	\\ \hline
850	 	&500		&\textbf{2.2} $\pm$ \textbf{2.1}			&\textbf{2.3} $\pm$\textbf{2.1}    &\textbf{2.9} $\pm$ \textbf{2.9} 	&\textbf{0.924}	\\ \hline
\end{tabular}
\caption{Estimation errors and mean accuracy of the shoulder pose estimation on \textit{Pandora}}
\label{tab:res_crop}
\end{table}
\noindent As already mentioned, in real situations the driver head may be affected by severe occlusions caused by hands and objects such as smartphones, scarves, bottles and so on. For this reason, we have carried out a specific set of experiments to test the reliability of \textit{POSEidon} in presence of occlusions or missing data. We artificially applied the masks depicted in Figure \ref{fig:occlusion} to remove parts of the input frames and simulate occlusions. The corresponding performance of \textit{POSEidon} is shown in Table \ref{tab:occlusions}, which confirms the reliability of the system also in these cases. The absence of the upper body part of the head strongly impacts with the system performance, in particular for the estimation of the pitch angle. Similarly, the head part around the nose plays a crucial role in the pose estimation, as highlighted by the errors generated by the occlusion type (e).  \\
The network performing the shoulder pose estimation has been tested on \textit{Pandora} only, due to the lack of the corresponding annotations in the other datasets. Results are reported in Table \ref{tab:res_crop}, where different image crops are compared (Section \ref{sec:shoulders}). The reported results are very promising, reaching an accuracy over $92\%$. \\
In order to have a fair comparison, results reported in Tables \ref{tab:resBiwiPose} and \ref{tab:resOurDataset} are obtained using the ground truth head position as input to the crop procedure. We finally test the whole pipeline, including the head localization  network described in section \ref{sec:headlocalization}, using also \textit{ICT-3DHP} dataset. The mean error of the head localization (in pixels) and the pose estimation errors are summarized in Table \ref{tab:pipeline}. Sometimes, the estimated position generates a more effective crop of the head. As a result, the whole pipeline performs better on the head pose estimation over the \textit{Biwi} dataset. \textit{POSEidon} reaches valuable results also on the \textit{ICT-3DHP} dataset and it provides comparable results with respect to state of the art methods working on both depth and RGB data (4.9$\pm$5.3, 4.4$\pm$4.6, 5.1$\pm$5.4 \cite{saeed2015}, 7.06, 10.48, 6.90 \cite{baltruvsaitis2012}, for pitch, roll and yaw respectively).\\
The complete framework has been implemented and tested on a desktop computer equipped with a \textit{NVidia Quadro k2200} GPU board and on a laptop with a \textit{NVidia GTX 860M}, exploiting \textit{Keras} \cite{chollet2015} with \textit{Theano} \cite{al2016theano} backend. Real time performance has been obtained in both cases, with a processing rate of more than 30 frames per second, with a limited dedicated graphical memory requirement. Some examples of the system output are reported in Figure \ref{fig:final demo}, where the six pose angles are visually shown using colored bars. In addition, the original depth map, the \textit{Face-from-Depth} reconstruction and the motion data given in input to \textit{POSEidon} are placed on the left of each frame. 
Pre-trained networks and models are publicly available.



\begin{table}[t!]
\centering
\small
\begin{tabular}{ccccc}
\hline
\textbf{Dataset} 	&\textbf{Loc.}  &\multicolumn{3}{c}{\textbf{Head}} \\ 
 					&	&Pitch			&Roll			&Yaw     			\\ \hline \hline
Biwi 				&3.27$\pm$2.19	& 1.5$\pm$1.4 	&1.7$\pm$1.7    &2.3$\pm$2.1   		\\ \hline
ICT-3DHP 			&-	& 5.0$\pm$4.3 	&3.5$\pm$3.5    &7.1$\pm$6.1  		\\ \hline
Pandora 			&4.27$\pm$3.25	& 7.6$\pm$8.5  	&4.8$\pm$4.8 	&10.6$\pm$12.7 	 	 	\\ \hline
\end{tabular}
\caption{Results on \textit{Biwi}, \textit{ICT-3DHP} and \textit{Pandora} dataset of the complete \textit{POSEidon} pipeline (\textit{i.e.}, head localization, cropping and pose estimation).}
\label{tab:pipeline}
\end{table}

\section{Conclusions and future work}
A complete framework for head localization and driver pose estimation called \textit{POSEidon} is presented. No previous computation of specific facial features is required. The system has shown real time and impressive results also in presence of occlusions, extreme poses of head and shoulders. Besides, the use of only depth data enhances the efficacy under different illumination conditions. 
All these aspects make the proposed framework suitable to particular challenging contexts, such as automotive. A new and high quality 3D dataset, \textit{Pandora}, is then proposed and publicly released. 
The system has been developed with a modular architecture: if it is possible to capture both RGB and depth images during the training, the complete architecture can be used. Otherwise, the \textit{Face-from-Depth} module should be removed from the system, using the depth+MI combination, reaching worst but still satisfactory performances. 

{\small
\bibliographystyle{ieee}
\bibliography{bibliography}
}


\end{document}